\pdfoutput=1

\documentclass[11pt]{article}

\usepackage{EACL2023}

\usepackage{times}
\usepackage{latexsym}
\usepackage{graphicx}
\usepackage{subcaption}
\usepackage[T1]{fontenc}
\usepackage{multirow}
\usepackage{amsmath}
\usepackage{amsfonts}
\usepackage{algorithm}
\usepackage{algpseudocode}
\usepackage[utf8]{inputenc}
\usepackage{todonotes}
\usepackage{microtype}

\usepackage{inconsolata}
\usepackage{array, caption, tabularx,  ragged2e,  booktabs}

\title{Model-based Counterfactual Generator for Gender Bias Mitigation}

\author{Ewoenam Kwaku Tokpo  \and Toon Calders \\
        Department of Computer Science, University of Antwerp, Belgium}

\begin{document}
\maketitle
\begin{abstract}
Counterfactual Data Augmentation (CDA) has been one of the preferred techniques for mitigating gender bias in natural language models. CDA techniques have mostly employed word substitution based on dictionaries.
Although such dictionary-based CDA techniques have been shown to significantly improve the mitigation of gender bias, in this paper, we highlight some limitations of such dictionary-based counterfactual data augmentation techniques, such as susceptibility to ungrammatical compositions, and lack of generalization outside the set of predefined dictionary words. 
Model-based solutions can alleviate these problems, yet the lack of qualitative parallel training data hinders development in this direction.
Therefore, we propose a combination of data processing techniques and a bi-objective training regime to develop a model-based solution for generating counterfactuals to mitigate gender bias.
We implemented our proposed solution and performed an empirical evaluation which shows how our model alleviates the shortcomings of dictionary-based solutions. 
\end{abstract}

\section{Introduction}
\label{sec: intro}

Language models (LMs) have primarily been trained on internet data, which has been shown to contain societal biases and stereotypes \cite{wolf2017we}, predisposing these models to perpetuate, and in some cases amplify these biases.
With the growing adoption of LMs and LM-based tools for a myriad of uses, it has become essential to mitigate these biases.
Efforts have been made, especially in the area of gender, to develop mitigation strategies to alleviate the effects of these biases.
One of the widely adopted techniques for mitigating gender bias has been counterfactual data augmentation (CDA).
The approach is intuitive;
for every text instance, explicit gender words are replaced with their equivalents from other gender groups. This ensures a balanced association between gender words and certain target words whose neutrality in association must be preserved in a given domain.
Using CDA to mitigate gender bias has been shown to produce significant improvements in practice \cite{hall-maudslay-etal-2019-name, tokpo2023far}.

Key works introducing CDA as a bias mitigation technique such as \citep{zhao-etal-2018-gender, lu2020gender, zmigrod-etal-2019-counterfactual} all adopt a word substitution approach based on dictionaries.
The process involves compiling -- usually manually -- a set of gender words and their opposite gender equivalents in a pairwise fashion.
A duplicate of each text instance is created, in which any occurrence of a gender-specific term from the dictionary is substituted with its respective counterfactual equivalent. This is to establish a balanced distribution of gender-related words within the text.

Although generative language models like GPT \cite{radford2018improving} have surged in popularity, dictionary-based techniques for counterfactual data generation continue to be the widely adopted approach.
This is primarily due to the relative unavailability of parallel texts required to effectively train such generative models. This has been a bottleneck in the development of model-based methods.


While dictionary-based word substitution techniques have been conventional to use, we will discuss some significant limitations associated with them in this paper.
We will highlight how such dictionary-based CDA methods are prone to generating grammatically incoherent instances.
Moreover, we will illustrate how these methods heavily depend on a limited set of terms, thereby curbing their ability to generalize beyond the terms within that list.

Consider these text fragments from the Bias-in-bios dataset \cite{de2019bias} \emph{\textbf{"Memory received her Bachelor and Masters of Accountancy..."}} and \emph{\textbf{"Laura discovered her passion for programming after teaching herself some Python..."}}, using a popular CDA technique by \citet{hall-maudslay-etal-2019-name}, generates \emph{\textbf{"Memory received his Spinster and Mistresses of Accountancy..."}} and 
\emph{\textbf{"Anthony discovered his passion for programming after teaching herself some Python..."}} respectively.
This is because the terms \emph{bachelor} and \emph{master} can have different meanings some of which could be gender (un)associated.
Secondly, the terms \emph{herself} and \emph{himself} are left out of the dictionary by \citet{hall-maudslay-etal-2019-name}. 

To resolve these issues, we introduce a model-based counterfactual generator as a CDA approach. 
We make two significant contributions in this regard. First, we propose a pipeline of techniques
to generate parallel data from dictionary-based CDA outputs.
Secondly, we introduce a bi-objective training approach to train a counterfactual generator. Our model is made up of a generator and a discriminator akin to adversarial models but with some nuances.
One key component of our bi-objective training approach is how we circumvent the bottleneck of backpropagating through a discrete layer of non-differentiable tokens, which is a common issue in textual models.  
A bi-objective approach also makes our model more robust to errors in the parallel data labels.
We will show how these techniques combine to solve the issues with dictionary-based methods.

To sum up our key contributions:
\begin{enumerate}
    \item  We propose a data processing pipeline to generate parallel data for training CDA models.
    \item  We propose a bi-objective model to generate counterfactuals and introduce a unique adaptation that allows the discriminator to circumvent backpropagating through a discrete layer of non-differentiable tokens.
\end{enumerate}

We make a version of the trained models and code publicly available\footnote{https://github.com/EwoeT/MBCDA} to facilitate further studies and improvements.

\section{Background}
\subsection{Related Literature}
Early works on CDA used simple rule-based approaches for data augmentation. They created dictionaries of gender terms and used matching rules to swap words \cite{de2019bias}. 
Later works began to incorporate grammatical information like part-of-speech tags to swap words \cite{zhao-etal-2018-gender}. 
Even with such improvements, there remained limitations such as the inability to generate counterfactuals for named entities like human names. Generating counterfactuals without proper name interventions could result in semantically incorrect sentences \cite{lu2020gender}.

To address this, \citet{lu2020gender} do not augment sentences or text instances containing proper nouns and named entities.
\citet{hall-maudslay-etal-2019-name} discuss how this approach could be less effective as names could serve as a proxy. 
\citet{zhao-etal-2018-gender} addressed this by anonymizing named entities by replacing them with special tokens. Albeit an improvement, this could be problematic since in real-world data, names are not anonymized, hence gender correlations are still maintained when deployed on real-world texts.

Lamenting on the aforementioned lack of parallel corpus for training neural models, \citet{zmigrod-etal-2019-counterfactual} used a series of unsupervised techniques such as dependency trees, lemmata, part-of-speech tags, and morpho-syntactic tags for counterfactual generation. Their technique also expanded CDA outside of the English language setting to languages with rich morphology.
However, their approach remains susceptible to ungrammatical compositions in the English language and is also unable to resolve named entities.

\citet{hall-maudslay-etal-2019-name} proposed a variant of CDA called counterfactual data substitution that introduced two key improvements. First, they introduced \emph{names intervention} to resolve the challenges of generating counterfactuals for named entities. They achieve this using a bipartite graph to match first names.
This intervention uses name frequency and degree of gender specificity based on data from the US social security Administration to generate a list of name pairs which is subsequently used to augment the original dictionary from \citet{lu2020gender}.
Secondly, they note that duplicating texts as is done in CDA could lead to unnatural statistical properties of texts like type-token ratios which could violate Heap's law, although this assumption was not empirically proven. 
Based on this postulation, they introduced the second improvement, \emph{data substitution}.
This seeks to preserve such statistical properties of texts by replacing about 50\% of the text instances with their counterfactuals instead of augmenting them. 
This non-duplication of texts is believed to improve the naturalness of the text.

\subsection{Lingering problems with CDA approaches}
Although CDA techniques have significantly evolved, two major limitations remain:

\paragraph{a. Dictionary restrictions:}
The chief problem with dictionary-based techniques is their inability to generalize beyond the dictionary word lists. 
As these dictionaries are manually compiled, they are in no way exhaustive.
This causes these techniques to be restrictive.

\paragraph{b. Ungrammatical compositions:}
\label{sec: ungrammatical}
Considerable efforts have been made in previous techniques to resolve ungrammatical counterfactual generations -- mostly addressed with part-of-speech tags. However, we still observe a persistent generation of ungrammatical texts as these techniques are non-contextualized and do not include context information of the text. 
We identify two main types of errors: 
\begin{enumerate}
    \item Errors resulting from improper/out-of-context substitutions: This is seen in our first example from section~\ref{sec: intro}, 
\emph{\textbf{"Memory received her Bachelor and Masters of Accountancy..."}} --> \emph{\textbf{"Memory received his Spinster and Mistresses of Accountancy..."}}.

\item Errors resulting from out-of-dictionary words: This is also manifested in our second example from section~\ref{sec: intro}. Since \emph{(herself, himself)} are not contained in the dictionary from \cite{hall-maudslay-etal-2019-name}, we encounter instances such as
\textbf{"Laura discovered her passion for programming after teaching \emph{herself} some Python..."} leading to
\textbf{"Anthony discovered his passion for programming after teaching \emph{herself} some Python..."}.
\end{enumerate}

\section{A model-based approach to CDA}

\begin{figure*}[tbh]
     \centering
     \begin{subfigure}[b]{0.49\textwidth}
         \centering
         \includegraphics[width=\textwidth]{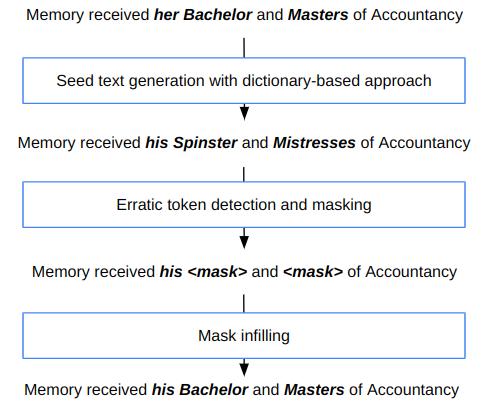}
         \caption{Parallel data generation -- Showing the pipeline of text transformations to generate parallel training data.}
         \label{fig:parallel_data_generation}
     \end{subfigure}
     \hfill
     \begin{subfigure}[b]{0.49\textwidth}
         \centering
         \includegraphics[width=\textwidth]{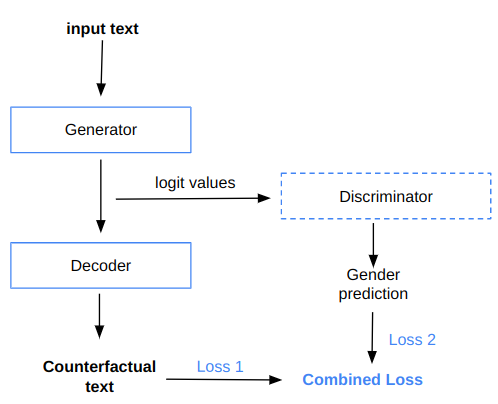}
         \caption{generator model -- Bi-objective model showing the use of a generator-discriminator set-up for training. Broken lines indicate the Discriminator is pretrained separately and frozen during the generator training.}
         \label{fig:generator_model}
     \end{subfigure}
      \caption{Our model-based approach. Figure (a) shows the data processing approach and Figure (b) shows the bi-objective training approach to train the generator.}
      \label{fig:model_arch}
\end{figure*}

The primary benefit of a model-based solution is that it allows the model to generalize outside the predefined list of seed words. The model learns latent attributes of gender words through which it can map gender words to their equivalents from another gender. As such, gender terms not explicitly included in the dictionary can be identified by the model via these latent features.

Implementing a model-based approach poses one major challenge: the unavailability of parallel data.
We thus propose a two-part solution for this. 
1) A data processing pipeline to transform dictionary-based model outputs into better-quality parallel data.
Since the counterfactual texts generated by these dictionary-based solutions can be error-riddled, directly using them as training data risks perpetuating these errors in the trained model.
2) A bi-objective training approach in which a discriminator constrains the generator during training to change the original gender of the text and further constrains it from predicting unlikely tokens that are frequently out-of-context. This further prevents the model from picking up lingering errors in the parallel labels.
We will expound on this in the subsequent sections and also examine the relevance of each of these steps in Section~\ref{subsec:ablation_analysis}.
These two processes are shown in Figure~\ref{fig:model_arch}.

\subsection{Parallel data generation}
\label{subsec: our_approach}
This step involves identifying and correcting grammatical inconsistencies from dictionary-based counterfactuals. We break this process down into four steps. Figure~\ref{fig:parallel_data_generation} illustrates this whole process.

\paragraph{Seed text generation:}
This initial step involves utilizing a dictionary-based technique to create seed texts, which will then undergo a transformation to parallel text labels for training.
For all instances in the training data, we generate counterfactuals using CDA with named intervention by \citet{hall-maudslay-etal-2019-name}. 

\paragraph{Erratic token detection:}
The idea here is to detect and mask tokens that have a  low probability of appearing in the context of a given text; following $t_i = t_{<mask>}$ if $P(t_i|T)<\theta$, where $T$ is the sequence of tokens, $t_i$ is the $i$th token in $T$, and $\theta$ is a predefined threshold value \footnote{We constantly use ELECTRA's default threshold logit value of 0.}. We define the resulting masked text as $T_\Pi$.
This is achieved using a pretrained ELECTRA model \cite{clark2020electra}. ELECTRA is an LM pretrained using a text corruption scheme -- text instances are corrupted by randomly replacing a number of tokens with plausible alternatives from BERT. ELECTRA is then trained to predict which tokens are real and fictitious.

Since the use of wordpiece tokenization causes issues as a word can be broken down into multiple subtokens, if a subtoken is selected for masking, we replace the entire sequence of associated subtokens with a {\tt<mask>} token.
For instance, \emph{``The men are duchesses''}, in a wordpiece tokenization could be decomposed to \emph{[``The'', ``men'', ``are'', ``duchess'', ``\#\#es'']}, Consequently, when \emph{``duchess''} is identified as an erratic token, the masking scheme replaces the entire subsequence \emph{[``duchess'', ``\#\#es'']}, thereby, generating \emph{``The men are <mask>''}.

\paragraph{Text correction with BART:}
Having obtained our masked intermediary texts, we generate plausible token replacements for each masked token. Since a {\tt<mask>} token could correspond to multiple subword tokens (as discussed above), the replacement generator should be capable of generating multiple tokens for a single {\tt<mask>} instance, making it suitable to use a generative model -- BART\cite{lewis-etal-2020-bart} -- to predict these replacement tokens. 
Because Masked Language Modeling is part of BART's pretraining objectives, we can utilize it in its pretrained form without the need for finetuning.
Given $T_\Pi$ from the previous step, the BART model tries to predict the correct filling for the masked token using the context $T_\Pi \backslash t_{<mask>}$, thus, the infilling task follows $P(t_{<mask>}|T_\Pi \backslash t_{<mask>})$.

\paragraph{Text filtration with BERT:}
Finally, we use a BERT classifier trained to detect gender to filter out resultant counterfactual text instances whose gender association remained the same as the original text.

\subsection{Bi-objective training}
Our training set-up consists of a generator and a discriminator.
We use a bi-objective approach to train the generator to maintain accurate text generation while ensuring a change in the gender association of the given text. This is illustrated in Figure~\ref{fig:generator_model}.

The primary goal of the generator model is to translate an input source text correctly into the target text. 
The discriminator model on the other hand ensures that the gender of the target text is changed.
The losses from both models (objectives) are combined in the loss defined as:
\begin{equation}
 {Loss} = \mathcal{L}_{generator} + \mathcal{L}_{discriminator}
\end{equation}

\paragraph{Generator model}
The generator  $h_{1_{\theta}}(X)$ is a BART model \cite{lewis-etal-2020-bart} trained with the parallel data obtained (section~\ref{subsec: our_approach}). The generator takes the original text as input and is trained to autoregressively generate the counterfactual of the source text using the corresponding parallel counterfactual texts as labels in a teacher-forcing manner \citep{williamsteacher}. We formulate this as:
\begin{equation}
 \mathcal{L}_{generator} = -\sum\limits_{t=1 }^klogP(y_t|Y_{<t}, X)   
\end{equation}
Where  $X$ and $Y$ are the source and target texts respectively, $y_t \in Y$ is the $t^{th}$ token in the target text and $Y_{<t}$ refers to all tokens in $y$ preceding $y_t$.

\paragraph{Discriminator}
The discriminator constrains the output text $Y$ from the generator to have the gender association flipped. The discriminator is separately pretrained to predict gender. When training the generator, the discriminator is frozen such that its weights are not updated. The discriminator loss is computed using the counterfactual gender label $\Bar{l_g} = 1-l_g$, where $l_g$ is the original gender label.

Training and deploying the discriminator on discrete tokens poses a major challenge, being: backpropagating through a non-differentiable discrete layer.
Some methods such as \cite{prabhumoye2018style, tokpo-calders-2022-text} propose using straight-through soft-sampling techniques to make one-hot encoding approximations using softmax values. 
We found this solution less effective. 
One alternative approach is to train and deploy the discriminator on sentence embeddings (hidden states) from the generator model instead of decoded tokens. However, the feature space of the resulting sentence embeddings changes as the model gets updated, making the discriminator insensitive to new representations.
We resolve this issue by training the discriminator $h_{2_{\theta}}(h_{1_{\theta}}(X))$ to predict gender based on logit values $h_{1_{\theta}}(X)$ from the generator instead of the text decoded from the argmax of the logits. 
This approach differs from soft-sampling straight-through approximations in the sense that, the discriminator is trained directly on logit values instead of one-hot encoding or discrete token inputs.
The rationale here is that logit values also capture information on non-argmax values, and also eliminate the need for approximations of one-hot encodings. Secondly, since these logits correspond directly to tokens in the vocabulary instead of latent features (as is the case when using sentence embeddings) they remain in the same feature space even as the generator model gets updated. Thus, the sensitivity of the discriminator is not affected. 

The discriminator is made up of a three-layer feed-forward network which takes the logit values of all tokens $y \in Y$ as input and outputs hidden states $Z \in \mathbb{R}^{|Y|\times d}$, where $d$ is the dimensionality of each tokens representation. 
$Z$ is mean-pooled to generate a single representative $\Bar{z}$ for $Y$. $\Bar{z}$ is fed through a one-layer feedforward network to predict the corresponding gender,
for which the loss is given as:
\begin{equation}
\mathcal{L}_{discriminator}=-logP(\Bar{g}|\Bar{z})
\end{equation}

\section{Experimental set-up}
To quantify and compare the effectiveness of our approach, we have carried out a series of experiments. The results of these evaluations will be discussed in section~\ref{sec: results}.

\subsection{Dataset}
We use data from wikidump \footnote{https://dumps.wikimedia.org/backup-index.html} to train the model.
For performance evaluation and other probes, we use the Bias-in-Bios dataset \cite{de2019bias} and the Jigsaw dataset \footnote{https://www.kaggle.com/c/Jigsaw-unintended-bias-in-toxicity-classification/data}.

\paragraph{Gender and target words} 
We compiled a list of gender and target words/terms from \citet{hall-maudslay-etal-2019-name, caliskan2017semantics, bolukbasi2016man}. 
We define gender words as explicit gender-identifying words like \emph{woman} and \emph{man} and target words as stereotyped words associated with corresponding gender groups, for example, \emph{nurse} and \emph{engineer}; in the same manner as \citet{caliskan2017semantics}.
We included the word list and its details in appendix~\ref{sec: wordlist}.

\subsection{Mitigation techniques}
For a comparative evaluation, we compare our model to \emph{CDA with names intervention} proposed by \citet{hall-maudslay-etal-2019-name} as this has been widely adopted and used in literature. We will refer to this as \textbf{CDA1}.
We also use \emph{CDA without names intervention}, which we will refer to as \textbf{CDA2}.
Furthermore, for the Bias-in-Bios dataset, we use the \textbf{Gender swapping} text that is the default counterfactual text accompanying the dataset. Text from \emph{Gender swapping} appear to have been generated using a simple replacement scheme without part-of-speech or name interventions and uses a limited set of words relative to \citet{hall-maudslay-etal-2019-name}.

\subsection{Evaluation metrics}
We evaluate our model in a multifaceted manner, looking at various evaluation metrics to quantify the performance of our approach from various perspectives. We have evaluated our work on \emph{fluency}, \emph{gender transfer accuracy}, \emph{extrinsic bias mitigation}, and \emph{intrinsic bias mitigation}. We further carried out ablation studies to see the effects of the various components of our approach and discuss some qualitative aspects of our work.

We will generally define bias as \emph{``the undesired variation of the [association] of certain words in that language model according to certain prior words in a given domain''} \citep{garrido2021survey}.

\section{Evaluation and results} 
\label{sec: results}

\subsection{Grammatical correctness and fluency}

We evaluate the model for grammatical coherence and fluency by computing the perplexity. 
Perplexity here refers to how well the text conforms to the probabilistic distribution of natural text as learned by a given pretrained language model.
With the definition $P(x) = \Pi^n_{i=1}P(s_n|s_1,...,s_{n-1})$, we use a distilled version\footnote{\url{https://huggingface.co/distilgpt2}} of  GPT-2 \cite{radford2019language} to compute the perplexity of the generated texts.

\begin{table}[tbh]
\centering
{
\begin{tabular}{lcc}
\hline
\textbf{} & \textbf{BiasinBios $\downarrow$} & \textbf{Jigsaw $\downarrow$}\\
\hline
Original* & 56.611& 100.787   \\
swapped & 60.813 & -- \\
CDA1 & 59.810 & 109.521 \\
CDA2 & 72.786 & 104.566 \\
MBCDA & \textbf{59.742} & \textbf{92.778} \\ \hline
\end{tabular}
}
\caption{
PPL of generated text using various CDA techniques. Lower scores indicate better fluency. -- The Jigsaw dataset does not include swapped texts.}
\label{tab: ppl_table}
\end{table}

In Table~\ref{tab: ppl_table}, the perplexity results show how well our model generates fluent text. Since many of the grammatical incoherents in the text are successfully resolved, the generated texts tend to be more fluent compared to the other CDA techniques.
This is consistently the case for both the Bias-in-Bios and the Jigsaw datasets.

For the original text of the Jigsaw dataset, it is surprising to see the high perplexity, since it is human-generated. However,we realized a number of grammatical and typographical mistakes that could account for this high perplexity.

\subsection{Gender transfer accuracy}
Here, we compute the percentage of texts that were flipped from the source gender to the target gender (the complementary gender group(s)); similar to the evaluation technique used for text style transfer by \citet{tokpo-calders-2022-text}.
We train a BERT model to predict the gender of the text.
With this, we quantify gender transfer accuracy as $1-probability\_of\_original\_gender$; hence, the low percentages of the original text.

As shown in Table ~\ref{tab: gen_trans_table}, our MBCDA model proved most effective in gender transfer when compared to all the other approaches for the Jigsaw dataset.
Also, MCBDA outperforms CDA1 and CDA2 on the Bias-in-bios dataset.

\begin{table}[tbh]
\centering
{
\begin{tabular}{lcc}
\hline
&{\textbf{BiasinBios \% $\uparrow$} } & {\textbf{\textbf{Jigsaw \%$\uparrow$}}}\\ \hline


Original* & 0.0 & 21.300  \\
swapped &  \textbf{99.100}  & --  \\
CDA1 & 98.710  &  75.702   \\
CDA2 & 98.424 & 74.662 \\
MBCDA & 99.000  & \textbf{76.300} \\ \hline
\end{tabular}
}
\caption{Gender transfer accuracy of the various CDA interventions. This indicates the percentage of counterfactual instances that were correctly converted to the complementary gender. \emph{The original samples have very low accuracies because no gender flipping occurs (original gender is preserved)}}.
\label{tab: gen_trans_table}
\end{table}

\subsection{Extrinsic bias mitigation}
We also test the resultant texts on actual downstream tasks, as this is usually the end-use of a language model.
We use downstream classification tasks corresponding to the respective datasets i.e. job prediction for Bias-in-bios and toxicity detection for the Jigsaw dataset. Using a trained BERT classifier, we compute the \emph{True Positive rate difference (TPRD)} and \emph{False Positive rate difference (FPRD)} between two gender groups as our measure of bias as was done by \citet{de2019bias}. 
$TPRD = P(\hat{y}=1|y=1,A=a) - P(\hat{y}=1|y=1,A=a^{\prime})$ and 
$FPRD = P(\hat{y}=1|y=0,A=a) - P(\hat{y}=1|y=0,A=a^{\prime})$.
Where $y$ is the true label, $\hat{y}$ is the predicted label, and $A$ is the gender group variable.
With this approach, we are able to check the efficacy of our approach in contextualized language model settings \citep{delobelle-etal-2022-measuring, may-etal-2019-measuring} (we will later evaluate our approach intrinsically on static non-contextuaised word embeddings).
\begin{table}[tbh]
\centering
{
\begin{tabular}{lcc}
\hline
\textbf{} & \textbf{BiasinBios $\downarrow$} & \textbf{Jigsaw $\downarrow$}\\
\hline
Original* & 0.105 & 0.940 \\
swapped & 0.052 & -- \\
CDA1 & 0.045 & \textbf{0.021} \\
CDA2 & 0.065 & 0.047 \\
MBCDA & \textbf{0.043} & 0.039 \\ \hline
\end{tabular}
}
\caption{
TPRD -- True positive rate difference between male and female text instance}
\label{tab: tprd_table}
\end{table}

\begin{table}[tbh]
\centering
{
\begin{tabular}{lcc}
\hline
\textbf{} & \textbf{BiasinBios $\downarrow$} & \textbf{Jigsaw $\downarrow$}\\
\hline
Original* & 0.107 & 0.937 \\
swapped & 0.054 & -- \\
CDA1 & 0.050 & \textbf{0.126} \\
CDA2 & 0.058 & 0.134\\
MBCDA & \textbf{0.043} & 0.161 \\ \hline
\end{tabular}
}
\caption{
FPRD -- False positive rate difference between male and female text instance}
\label{tab: fprd_table}
\end{table}

As shown in Table~\ref{tab: tprd_table} and Table~\ref{tab: fprd_table}, our model outperforms all the other approaches in both TPRD and FPRD for the Bias-in-Bios dataset.
CDA1 on the other hand showed better mitigation effects for the Jigsaw dataset for both TPRD and FPRD.

\subsection{Intrinsic bias mitigation}

We train a word2vec \cite{mikolov2013efficient} model with the different CDA techniques.
We then use the Word Embedding Association Test (WEAT) \cite{caliskan2017semantics} which measures the association between gender words and target words.

\[
s({X}, {Y},{A}, {B}) = \sum_{x \in {X}} u(x,{A}, {B}) -\sum_{y \in {Y}} u(y,{A}, {B})
\]

\[
s(w,A,B) = \underset{a \in A}{mean} \: cos(\vec{w},\vec{a})- \underset{b \in B}{mean} \:  cos(\vec{w},\vec{b})
\]

Where $X$ and $Y$ are female and male sets of target words respectively, $A$ and $B$ are female and male sets of attribute words respectively and $cos()$ is the cosine similarity function.
We use gender and target terms from \citet{caliskan2017semantics}. Specifically, we use career words for the Bias-in-Bios dataset since it is focused on professions, and (un)pleasant terms from WEAT 1 for the jigsaw dataset since it relates to toxicity prediction. The exact wordlists are given in Appendix~\ref{sec: weat_wordlist}.

WEAT results in Table~\ref{tab: weat1} and Table~\ref{tab: weat2} illustrate how MBCDA improves intrinsic bias mitigation as per WEAT scores and effect size. The full results for other embedding sizes are given in Table ~\ref{tab: weat_bios_appendix} and Table~\ref{tab: weat_jigsaw_appendix} of Appendix~\ref{sec:appendix_weat}.

\begin{table}[tbh]
\centering
{
\begin{tabular}{lcc}
\hline
& \multicolumn{2}{c }{\textbf{\textbf{Bias-in-Bios}}}\\ \hline
 & \textbf{WEAT$\downarrow$} & \textbf{Effect\_size$\downarrow$} \\
\midrule
bios & 0.382 & 0.309 \\
swapped & 0.101 & 0.121 \\
CDA1 & 0.087 & \textbf{0.084}  \\
CDA2 & 0.080 & 0.092  \\
MBCDA & \textbf{-0.068} & -0.108 \\
\hline
\end{tabular}
}
\caption{WEAT Bias-in-Bios -- Embedding size of 100}
\label{tab: weat1}
\end{table}

\begin{table}[tbh]
\centering
{
\begin{tabular}{lcc}
\hline
& \multicolumn{2}{c }{\textbf{\textbf{Jigsaw}}}\\ \hline
 & \textbf{WEAT$\downarrow$} & \textbf{Effect\_size$\downarrow$} \\
\midrule
bios & 0.047 & 0.352\\
CDA1 &  0.035 & 0.043 \\
CDA2 &  0.081 & 0.102 \\
CDS & -0.021 & -0.042 \\
MBCDA & \textbf{-0.012} & \textbf{-0.015}\\
\hline
\end{tabular}
}
\caption{WEAT Jigsaw -- Embedding size of 100}
\label{tab: weat2}
\end{table}

\subsection{Qualitative analysis}
\begin{figure}[tbh]
     \centering
         \includegraphics[width=0.9\columnwidth]{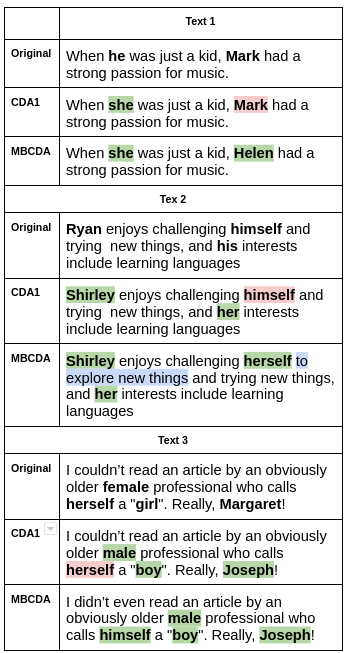}
      \caption{Text extracts from datasets. Green highlights indicate correct counterfactuals, red highlights indicate wrong counterfactuals and blue highlights indicate differences in generation.}
      \label{fig:text_samples_figure}
\end{figure}

A manual check of samples from both the Bias-in-Bios dataset and the Jigsaw datasets shows how effectively our model-based approach rectifies the issues with dictionary-based methods discussed in this paper. Figure~\ref{fig:text_samples_figure} shows three instances that highlight the effectiveness of our approach. 
In Text 1, we see MBCDA being able to properly resolve the name which is absent from the dictionary \citep{hall-maudslay-etal-2019-name}. This shows how well the model is able to learn gender attributes and generalize outside the given set of keywords in the dictionary model.
Text 2 and Text 3 also depict this ability. The words "\emph{himself}" and "\emph{herself}" that were also left out of the dictionary by \citet{hall-maudslay-etal-2019-name} were properly resolved.
In the majority of instances where the dictionary-based instances depicted such errors, our MBCDA model successfully resolved this.
We give a more extensive set of examples in Table~\ref{tab: samples_bios} and Table~\ref{tab: samples_jigsaw} in Appendix~\ref{sec:appendix_text_samples}, and include the complete output files for all methods and both datasets in the supplementary materials -- We draw readers' attention to offensive texts in the Jigsaw dataset.

\subsection{Ablation analysis}
\label{subsec:ablation_analysis}
Finally, we carried out ablation studies to see how the model performed when various components were taken out. We tested for six versions of MBCDA:
\begin{itemize}
    \item MBCDA1: default MBCDA model.
    \item MBCDA2: MBCDA trained without a discriminator
    \item MBCDA3:MBCDA trained on unfiltered text
    \item MBCDA4: MBCDA trained on unfiltered text and no discriminator
    \item MBCDA5: MBCDA trained directly on dictionary-based output without test processing
    \item MBCDA6: MBCDA trained directly on dictionary-based output without test processing and without a discriminatory.
\end{itemize}

In Table~\ref{tab: ablation2} and Table~\ref{tab: ablation3}, we show that training without a discriminator generally increases the fluency of the generated text but reduces the transfer accuracy. We also realize that as the quality of the parallel label decreases, especially in the case of MBCDA5 and MBCDA6, the overall performance of the model deteriorates.

\begin{table*}[tbh]
\centering
{
\begin{tabular}{lcccccc}
\hline
\textbf{} & \textbf{PPL$\downarrow$} & \textbf{T-Acc$\uparrow$} & \textbf{TPRD$\downarrow$} & \textbf{FPRD$\downarrow$} & \textbf{WEAT$\downarrow$} & \textbf{Effect\_size$\downarrow$}\\
\hline
MBCDA1 & 59.742 & \textbf{99.01} & \textbf{0.043} & \textbf{0.043} & -0.068 & -0.108 \\
MBCDA2 & 59.192 & 98.521 & 0.052 & 0.045 & -0.068 & -0.078 \\
MBCDA3 & 60.934 & 98.972 & 0.055 & 0.044 & \textbf{0.005} & \textbf{0.006}  \\
MBCDA4 & \textbf{54.689}  & 98.591 & 0.054 & 0.047 & -0.221  & -0.233\\
MBCDA5 & 69.313 & 66.351 & 0.097 & 0.086 & 0.005  & 0.006\\
MBCDA6& 68.737 & 50.935 & 0.162 & 0.156 & -0.221  & -0.233 \\ \hline
\end{tabular}
}
\caption{Ablation study of MBCDA on Bias-in-Bios dataset}
\label{tab: ablation2}
\end{table*}

\begin{table*}[tbh]
\centering
{
\begin{tabular}{lcccccc}
\hline
\textbf{} & \textbf{PPL$\downarrow$} & \textbf{T-Acc$\uparrow$} & \textbf{TPRD$\downarrow$} & \textbf{FPRD$\downarrow$} & \textbf{WEAT$\downarrow$} & \textbf{Effect\_size$\downarrow$}\\
\hline
MBCDA1 & 92.778  & \textbf{76.358} & \textbf{0.039} & 0.161 & \textbf{-0.012} & \textbf{-0.015}   \\
MBCDA2 & 84.176 & 75.197 & 0.099 & 0.161 & 0.066 & 0.069   \\
MBCDA3 & 88.060 & 75.591 & 0.089 & 0.172 & 0.032  & 0.031 \\
MBCDA4 & \textbf{80.306} & 74.933 & 0.047 & \textbf{0.145} & 0.133 & 0.128 \\
MBCDA5 & 149.862 & 50.270 & 0.820 & 0.663 & 0.091 & 0.140   \\
MBCDA6 & 150.138 & 50.929 & 0.803 & 0.738 & 0.139 &  0.223 \\ \hline
\end{tabular}
}
\caption{Ablation study of MBCDA on Jigsaw dataset}
\label{tab: ablation3}
\end{table*}

\section{Conclusion}
In this paper, we highlight some issues that pertain to dictionary-based counterfactual data augmentation techniques. We discuss how these techniques are prone to grammatical incoherencies and lack generalization outside dictionary terms.
On that basis, we propose a model-based approach we call MBCDA.
This uses a combination of data processing and a bi-objective training approach to generate counterfactual texts for binary gender.
We show that this approach can resolve these challenges associated with dictionary-based methods. We also carry out evaluations to determine the mitigation effect of our approach. We find that it demonstrates better mitigation capability compared to dictionary-based approaches.

\section*{Limitations}
One observation during training was an apparent trade-off between our proposed model's effectiveness in learning latent gender attributes such that it could generalize outside the set of dictionary words and the susceptibility of the model to pick some of the ungrammatical patterns from the dictionary-based methods.
As such, although very few, we found instances where the model repeated some errors associated with the dictionary-based approaches. 

Secondly, this work does not include an extensive investigation of the effect of hyper-parameter choices on the performance of the model, leaving open the possibility of an even better performance of the model. Such investigations will be carried out in our future works.

Again, we were unable to investigate the applicability of the proposed solution to non-binary cases of bias. Even though adaptations could be made to that effect, it is challenging to confirm without empirical tests how effective that would be.

Finally, the absence of formal human evaluation by external evaluators limits more detailed and objective discussions from a qualitative perspective. We seek to include this in our upcoming works to inform more concise discussions.

\section*{Ethics Statement}

From an ethical perspective, when adopting this work in broader contexts, especially outside the English language domain, one should consider variations in societal stereotypes. Most of the data including gender terms, and stereotypes used in this work are from North American sources. Consequently, these conclusions may not necessarily generalize or hold in different language or cultural domains. The impact of this work may be different and unpredictable in those domains.

Secondly, since this work does not investigate the impact of the proposed solution in non-binary cases of bias, it is difficult to affirm its efficacy in such settings.

\section*{Acknowledgements}
Ewoenam Kwaku Tokpo received funding from the Flemish Government under the ``Onderzoeksprogramma Artificiële Intelligentie (AI) Vlaanderen'' programme.

\bibliography{anthology,custom}
\bibliographystyle{acl_natbib}

\appendix
\section*{Appendix}
\label{sec:appendix}

\section{Experimental setup}
We run our experiments on Tesla V100-SXM3 GPUs with 32GB memory
With the exception of the parameters below, we use all default parameters of the various pretrained language models we used.
\begin{itemize}
    \item Number of epochs for counterfactual generator model: 2
    \item Learning rate: $2e^{-5}$ 
    \item batch size: 48
    \item Number of epochs for counterfactual discriminator models: 4
    \item Accuracy of discriminator models: 1.0
\end{itemize}

BERT classifiers for gender prediction were all trained on the Bias-in-Bios dataset.

\section{Dataset}
For the Bias-in-Bios dataset, we convert the labels to binary, by categorizing all male-dominated jobs as one label and all female-dominated jobs as another label.
For both the Jigsaw and the Bias-in-Bios datasets, we create an imbalance in the dataset set by sampling instances such that certain genders were more heavily associated with certain labels. With this, we artificially introduce more bias in the original dataset so the mitigation effects of the various approaches can be more visibly detected. 
 
\paragraph{Bias-in-bios}
Dataset size for experiments size: 21499

 Gender-label distribution:
 \begin{itemize}
    \item Female instances for female-labeled-jobs: 8286
    \item Female instances for male-labeled-jobs: 1936
    \item Male instances for female-labeled-jobs: 2419
    \item Male instances for male-labeled-jobs: 8858
\end{itemize}

\paragraph{Jigsaw}
Dataset size for experiments: 5595

We labeled the gender of the data using the original numerical gender annotations (0 to 1). We selected $(female>0.5) \& (male<0.5)$ as female instances and $(female<0.5) \& (male>0.5)$ as male instances. 

 Gender-label distribution:
 \begin{itemize}
    \item Female instances for toxic labels: 2595
    \item Female instances for non-toxic labels: 1000
    \item Male instances for toxic labels: 1000
    \item Male instances for non-toxic labels: 3000
\end{itemize}

\section{Word lists}
\label{sec: wordlist}

\subsection{WEAT tests}
\label{sec: weat_wordlist}

\begin{itemize}
    \item \textbf{female list: }mother,
 girlfriend,
 mom,
 females,
 sister,
 grandmother,
 wife,
 mothers,
 her,
 women,
 herself,
 girl,
 female,
 daughter,
 girls,
 aunt,
 she,
 her,
 sisters,
 woman,
 bride,
 daughters

     \item \textbf{male list: }father,
 boyfriend,
 dad,
 males,
 brother,
 grandfather,
 husband,
 fathers,
 him,
 men,
 himself,
 boy,
 male,
 son,
 boys,
 uncle,
 he,
 his,
 brothers,
 man,
 groom,
 sons

\item \textbf{female career stereotypes: }server,
 therapist,
 secretary,
 teacher,
 psychologist,
 host,
 assistant,
 pathologist,
 model,
 dietitian,
 nurse

\item  \textbf{male career stereotypes: }
chef,
clergy,
architect,
pilot,
conductor,
firefighter,
engineer,
painter,
programmer

\end{itemize}

\begin{itemize}
\item \textbf{Pleasant words: } caress, freedom, health, love, peace, cheer, friend, heaven, loyal, pleasure, diamond, gentle, honest, lucky,
rainbow, diploma, gift, honor, miracle, sunrise, family, happy, laughter, paradise, vacation.

\item  \textbf{Unpleasant words: }
abuse, crash, filth, murder, sickness, accident, death, grief, poison, stink, assault, disaster, hatred, pollute,
tragedy, divorce, jail, poverty, ugly, cancer, kill, rotten, vomit, agony, prison.

\end{itemize}

\subsection{CDA}
For counterfactual data augmentation, we use the dictionary from \citet{hall-maudslay-etal-2019-name}

\section{Text samples}
\label{sec:appendix_text_samples}

Text instances for Bias-in-Bios are given in Table~\ref{tab: samples_bios} and text instances for the Jigsaw dataset are given in ~\ref{tab: samples_jigsaw}. -- We draw readers' attention to offensive texts in the Jigsaw dataset.

\begin{table*}[tbh]
\noindent\setlength\tabcolsep{4pt}%
\begin{tabularx}{\linewidth}{|l|*{3}{>{\RaggedRight\arraybackslash}X|}}
\hline
\textbf{\#} &\textbf{Original} & \textbf{Dictionary-based (CDA)} & \textbf{MBCDA}\\
\hline
\midrule
358 & \emph{\textbf{Amy}} received \emph{\textbf{her}} \emph{\textbf{Bachelor}} of Science in animal science from North Carolina State University in 2007 and \emph{\textbf{her}} \emph{\textbf{Bachelor}} of Nursing from the University of North Carolina in 2009. & 
\emph{\textbf{Frank}} received \emph{\textbf{his}} \emph{\textbf{Spinster}} of Science in animal science from North Carolina State University in 2007 and \emph{\textbf{his}} \emph{\textbf{Spinster}} of Nursing from the University of North Carolina in 2009.&
\emph{\textbf{Frank}} received \emph{\textbf{his}} \emph{\textbf{Bachelor}} of Science in Animal Science from North Carolina State University in 2007 and \emph{\textbf{his}} \emph{\textbf{Master}} of Nursing from the University of North Carolina in 2009.
\\ \hline
369 & When \emph{\textbf{he}} was just a kid, \emph{\textbf{Mark}} had a strong passion for music. \emph{\textbf{He}} loved it so much that \emph{\textbf{he}} decided he would love to create music \emph{\textbf{himself}} and also become a DJ.&
 When \emph{\textbf{she}} was just a kid, \emph{\textbf{Mark}} had a strong passion for music. S\emph{\textbf{he}} loved it so much that \emph{\textbf{she}} decided \emph{\textbf{she}} would love to create music \emph{\textbf{himself}} and also become a DJ. &
 When \emph{\textbf{she}} was just a kid, \emph{\textbf{Helen}} had a strong passion for music. \emph{\textbf{She}} loved it so much that \emph{\textbf{she}} decided \emph{\textbf{she}} would love to create music and also become a DJ.
 \\ \hline
238 & \emph{\textbf{He}} has over 11 years experience in the industry and prides \emph{\textbf{himself}} on \emph{\textbf{his}} hands-on approach...&
 \emph{\textbf{She}} has over 11 years experience in the industry and prides \emph{\textbf{himself}} on \emph{\textbf{her}} hands-on approach... &
 \emph{\textbf{She}} has over 11 years experience in the industry and prides \emph{\textbf{herself}} on \emph{\textbf{her}} hands-on approach...
 \\ \hline
 198 & \emph{\textbf{She}} has been living in Tokyo since 2006. \emph{\textbf{She}} received \emph{\textbf{her}} BFA in Design at Cornish College of the Arts in Seattle, followed by \emph{\textbf{her}} \emph{\textbf{Masters}} of Architecture at the University of Oregon.&
198 \emph{\textbf{He}} has been living in Tokyo since 2006. He received his BFA in Design at Cornish College of the Arts in Seattle, followed by \emph{\textbf{his}} \emph{\textbf{Mistresses}} of Architecture at the University of Oregon.&
\emph{\textbf{He}} has been living in Tokyo since 2006. \emph{\textbf{He}} received his BFA in design at Cornish College of the Arts in Seattle, followed by \emph{\textbf{his}} \emph{\textbf{Masters}} of Architecture at the University of Oregon. \\ \hline
 3 & \emph{\textbf{His}} children's books include "Pouch Potato, \emph{\textbf{Charlie}} and the Caterpillar, \emph{\textbf{King Bob's}} New Clothes," and numerous cookbooks including "Eat This" and "Eat This Again." See less&
\emph{\textbf{Her}} children's books include "Pouch Potato, \emph{\textbf{Shannon}} and the Caterpillar, \emph{\textbf{Queen Lula's}} New Clothes," and numerous cookbooks including "Eat This" and "Eat This Again." See less&
\emph{\textbf{Her}} children's books include "Pouch Potato, \emph{\textbf{Shannon}} and the Caterpillar, \emph{\textbf{Queen Lula's}} New Clothes," and numerous cookbooks including "Eat This" and "Eat That Again." See less \\ \hline
56 &
\emph{\textbf{Lorna}} is the Director of the Assertive Community Treatment (ACT) Technical Assistance within the Center for Excellence in Community Mental Health at the University of North Carolina at Chapel Hill. \emph{\textbf{Lorna's}} clinical training... &
\emph{\textbf{Tomas}} is the Director of the Assertive Community Treatment (ACT) Technical Assistance within the Center for Excellence in Community Mental Health at the University of North Carolina at Chapel Hill. \emph{\textbf{Tomas's}} clinical training&
\emph{\textbf{Brett}} is the director of the assertive community treatment (ACT) technical assistance within the Center for Excellence in Community Mental Health at the University of North Carolina at Chapel Hill. \emph{\textbf{Brett's}} clinical training... \\ 

\bottomrule
\end{tabularx}
\caption{Text samples -- Bias in Bios dataset}
\label{tab: samples_bios}
\end{table*}

\begin{table*}[tbh]
\noindent\setlength\tabcolsep{4pt}%
\begin{tabularx}{\linewidth}{|l|*{3}{>{\RaggedRight\arraybackslash}X|}}
\hline
\textbf{\#} &\textbf{Original} & \textbf{Dictionary-based (CDA)} & \textbf{MBCDA}\\
\hline
\midrule
57 & What's up with so called "Conservatives" and their lists?     Is this \emph{\textbf{man}}, who fancies \emph{\textbf{himself}} a "Libertarian" going to keep a file on everybody?&
What's up with so called "Conservatives" and their lists?     Is this \emph{\textbf{woman}}, who fancies \emph{\textbf{himself}} a "Libertarian" going to keep a file on everybody? &
What's up with so called \emph{"libertarians"} and their lists?  Is this \emph{\textbf{woman}}, who fancies \emph{\textbf{herself}} a "libertarian" going to keep a file on everybody? \\ \hline
281 & dumb people it proof \emph{\textbf{she}} does not care bout anyone bout \emph{\textbf{herself}} and \emph{\textbf{her}} money and proof she nacarssit blame others for \emph{\textbf{her}} loss...&
dumb people it proof \emph{\textbf{he}} does not care bout anyone bout \emph{\textbf{herself}} and \emph{\textbf{his}} money and proof \emph{\textbf{he}} nacarssit blame others for \emph{\textbf{his}} loss...&
Dumb people it proof \emph{\textbf{he}} does not care bout anyone bout \emph{\textbf{himself}} and \emph{\textbf{his}} money and proof \emph{\textbf{he}} nacarssit blame others for \emph{\textbf{his}} loss... \\ \hline
296 & Stupid owner. Poor dog that belongs to \emph{\textbf{her}}... "Atta \emph{\textbf{girl}}!!" to the \emph{\textbf{woman}} who broke the window. &
Stupid owner. Poor dog that belongs to \emph{\textbf{him}}... "Atta \emph{\textbf{boy}}!!" to the \emph{\textbf{man}} who broke the window.&
Stupid owner. Poor dog that belongs to \emph{\textbf{him}}... "Atta \emph{\textbf{boy}}!!" to the \emph{\textbf{man}} who broke the window. \\ \hline
1347 & \emph{\textbf{Ms.}} Eddo-Lodge is just another Human who is looking for a cheap and easy 'ticket' to make money by using cheap sensationalism.. the reality is that the \emph{\textbf{woman}} has hit upon \emph{\textbf{her}} 'cheap \& easy' bigotry as a mission statement..&
\emph{\textbf{Ms.}} Eddo-Lodge is just another Human who is looking for a cheap and easy 'ticket' to make money by using cheap sensationalism.. the reality is that the \emph{\textbf{man}} has hit upon \emph{\textbf{his}} 'cheap \& easy' bigotry as a mission statement..& 
\emph{\textbf{Mr.}} Eddo-Lodge is just another human who is looking for a cheap and easy 'ticket' to make money by using cheap sensationalism.. The reality is that the \emph{\textbf{man}} has hit upon \emph{\textbf{his}} 'cheap \& easy' bigotry as a mission statement..\\ \hline
40 & The \emph{\textbf{male}} who was found crying in the hall discovered them.  \emph{\textbf{He}} returned to find them.  Video shows that.  \emph{\textbf{He}} has been cleared.  I know \emph{\textbf{him}}.&
The \emph{\textbf{female}} who was found crying in the hall discovered them.  \emph{\textbf{She}} returned to find them.  Video shows that.  \emph{\textbf{She}} has been cleared.  I know \emph{\textbf{her}}.&
The \emph{\textbf{female}} who was found crying in the hall discovered them. \emph{\textbf{She}} returned to find them. Video shows that. \emph{\textbf{She}} has been cleared. I know \emph{\textbf{her}}. \\ \hline
310 & The \emph{\textbf{feminists}} in the so-called \emph{\textbf{feminist}} movement are the epitome of penis envy.  They do not like who they are.  The result being they actually denigrate the truly \emph{\textbf{feminine}}. &
 The \emph{\textbf{feminists}} in the so-called \emph{\textbf{feminist}} movement are the epitome of penis envy.  They do not like who they are.  The result being they actually denigrate the truly \emph{\textbf{feminine}}. &
 The \emph{\textbf{feminists}} in the so-called \emph{\textbf{feminist}} movement are the epitome of penis envy. They do not like who they are. The result being they actually denigrate the truly \emph{\textbf{feminine}}.\\ 
\bottomrule
\end{tabularx}
\caption{Text samples -- Jigsaw dataset}
\label{tab: samples_jigsaw}
\end{table*}

\section{Full WEAT results}
\label{sec:appendix_weat}
Full results for all embedding sizes are shown in Table ~\ref{tab: weat_bios_appendix} for the Bias-in-Bios dataset and Table~\ref{tab: weat_jigsaw_appendix} for the Jigsaw dataset.

\begin{table*}[h]
\centering
{
\begin{tabular}{llrrrrrr}
\hline
& \multicolumn{2}{c}{{Embedding size: 50}} & \multicolumn{2}{c}{{Embedding size: 100}} & \multicolumn{2}{c}{{Embedding size: 300}} \\ \hline
 & weat & effect\_size & weat & effect\_size & weat & effect\_size \\
\midrule
bios & 0.190784 & 0.172840 & 0.381819 & 0.308984 & 0.375666 & 0.285945 \\
swapped & 0.156061 & 0.175832 & 0.100694 & 0.121057 & 0.102255 & 0.127662 \\
CDA & 0.068901 & 0.059151 & 0.086736 & 0.083767 & 0.041385 & 0.037845 \\
CDA2 & 0.051203 & 0.064703 & 0.079712 & 0.092326 & 0.123123 & 0.141099 \\
MBCDA1 & -0.069004 & -0.094638 & -0.068025 & -0.107624 & -0.057603 & -0.080094 \\
MBCDA2 & -0.042068 & -0.054377 & -0.067622 & -0.078216 & -0.050400 & -0.055534 \\
MBCDA3 & 0.033504 & 0.035228 & 0.005354 & 0.006447 & 0.014445 & 0.015819 \\
MBCDA4 & -0.274253 & -0.276636 & -0.221395 & -0.233077 & -0.248930 & -0.267835 \\
MBCDA5 & 0.033504 & 0.035228 & 0.005354 & 0.006447 & 0.014445 & 0.015819 \\
MBCDA6 & -0.274253 & -0.276636 & -0.221395 & -0.233077 & -0.248930 & -0.267835 \\
\bottomrule
\end{tabular}
}
\caption{WEAT Bias-in-Bios}
\label{tab: weat_bios_appendix}
\end{table*}

\begin{table*}[h]
\centering
{
\begin{tabular}{llrrrrrr}
\hline
& \multicolumn{2}{c}{{Embedding size: 50}} & \multicolumn{2}{c}{{Embedding size: 100}} & \multicolumn{2}{c}{{Embedding size: 300}} \\ \hline
 & weat & effect\_size & weat & effect\_size & weat & effect\_size \\
\midrule
bios & 0.034623 & 0.205174 & 0.047315 & 0.352058 & 0.040549 & 0.386313 \\
CDA1 & 0.030983 & 0.037171 & 0.035416 & 0.042942 & 0.052212 & 0.066718 \\
CDA2 & 0.083400 & 0.099742 & 0.081324 & 0.101642 & 0.066696 & 0.088542 \\
MBCDA1 & -0.000763 & -0.000999 & -0.012062 & -0.015413 & 0.005423 & 0.007439 \\
MBCDA2 & 0.051880 & 0.053184 & 0.065934 & 0.068701 & 0.065925 & 0.073041 \\
MBCDA3 & 0.042385 & 0.040633 & 0.032127 & 0.031200 & 0.037822 & 0.037291 \\
MBCDA4 & 0.130284 & 0.125437 & 0.133372 & 0.127552 & 0.125752 & 0.121292 \\
MBCDA5 & 0.090506 & 0.134429 & 0.091247 & 0.140653 & 0.093130 & 0.165794 \\
MBCDA6 & 0.139705 & 0.225433 & 0.138603 & 0.223437 & 0.135682 & 0.248827 \\
\bottomrule
\end{tabular}
}
\caption{WEAT Jigsaw}
\label{tab: weat_jigsaw_appendix}
\end{table*}

\end{document}